\documentclass[graybox]{styles/svmult}

\usepackage{mathptmx}       
\usepackage{helvet}         
\usepackage{courier}        
\usepackage{type1cm}        
\usepackage{makeidx}         
\usepackage{graphicx}        
\usepackage{multicol}        
\usepackage[bottom]{footmisc}
\usepackage{amsmath}

\makeindex             

\begin{document}

\title*{Investigating Reinforcement Learning for Communication Strategies in a
Task-Initiative Setting}
\titlerunning{Grounding System Contributions}
\author{Baber Khalid and Matthew Stone}
\institute{Baber Khalid \and Matthew Stone \at Rutgers University, 110 Frelinghuysen Rd.
Piscataway, NJ 08854-8019 , \email{baber.khalid@rutgers.edu; mdstone@cs.rutgers.edu}}
%
%
\maketitle

\abstract*{Many conversational domains require the system to present nuanced information to users. Such systems must follow up what they say to address clarification questions and repair misunderstandings. In this work, we explore this interactive strategy in a referential communication task. Using simulation, we analyze the communication trade-offs between initial presentation and subsequent followup as a function of user clarification strategy, and compare the performance of several baseline strategies to policies derived by reinforcement learning. We find surprising advantages to coherence-based representations of dialogue strategy, which bring minimal data requirements, explainable choices, and strong audit capabilities, but incur little loss in predicted outcomes across a wide range of user models. }

\abstract{Many conversational domains require the system to present nuanced information to users. Such systems must follow up what they say to address clarification questions and repair misunderstandings. In this work, we explore this interactive strategy in a referential communication task. Using simulation, we analyze the communication trade-offs between initial presentation and subsequent followup as a function of user clarification strategy, and compare the performance of several baseline strategies to policies derived by reinforcement learning. We find surprising advantages to coherence-based representations of dialogue strategy, which bring minimal data requirements, explainable choices, and strong audit capabilities, but incur little loss in predicted outcomes across a wide range of user models. }

\section{Introduction}

Task-oriented dialogue systems have robust policies to make sure the system correctly captures user-specified parameters \cite{DBLP:conf/aaai/Cohen20}, but task-oriented interactions can also include points where the system's contributions are essential, such as information presentation \cite{658969}, constraint satisfaction \cite{DBLP:conf/aaai/Cohen20}, and real-world coordination \cite{bohus2009open-world}. At such points, task success will typically require that the system work across turns to make sure that its contributions become common ground with users. To achieve common ground, the system may need to draw inferences about what the user understands based on what the user says and does \cite{stone:lascarides:2010}, and act preemptively to resolve misunderstanding. At the same time, the system can expect users to work collaboratively to confirm their own understanding \cite{CLARK1989259}. When they do so, the system must be able to play its part in users' grounding strategies. In fact, fielded systems rarely have such abilities---they typically cannot answer users' clarification questions, for example \cite{larsson-2017-user}---and the potential impacts of such dynamics on system dialogue strategy are not well studied. This paper highlights some of the complexities and differences of grounding system contributions as compared to grounding user contributions, the more commonly studied interaction.

As a case study, we explore the visual-linguistic colors in context (CIC) task \cite{monroe-etal-2017-colors}.  This is a situated referential communication task, similar to those long studied in the psycholinguistic literature \cite{CLARK1989259}.  One participant, the director, has the job of identifying a target (one of three color patches presented on a computer display) to the other participant, the matcher. We explore the trade-offs in state representation, dialogue strategy, and design methodology for interactions where the system has the role of the director.  Our problem raises several novel questions for dialogue system design: How important is it for the system to anticipate and defuse potential user confusion or misunderstanding in formulating its contributions? How important is it for the system to learn, infer, and respond to the user's specific uncertainty or difficulty when the user asks for clarification? What empirical factors might influence the effectiveness of different system strategies and representations? As Section~\ref{sec:related} explains, the novelty of our work is addressing grounding from the perspective of what the user understands.  Section~\ref{sec:approach} describes the task and dataset, explains its intuitive challenges, and reviews the technical infrastructure that enables our experiments.

To investigate the trade-offs presented to a system behaving as the director, we delveop communication policies in two ways:
\begin{itemize}
    \item In Section~\ref{sec:rl_setup}, we introduce a framework to optimize the director communication strategy using an RL  approach.
    \item In Section~\ref{sec:baseline}, we review and analyze human-human conversations to handcraft a simple rule-based communication strategy.
\end{itemize}
The goal of these two strategies is to empirically contrast performance between two types of director policies and use the obtained results to analyzes strengths and weakness of each approach. A shared feature of both is a coherence-based dialogue architecture \cite{khalid-etal-2020-discourse,stone:lascarides:2010} that naturally provides the capability to answer task-specific clarification questions.

Section~\ref{sec:expt} compares the behavior and predicted results of our two approaches in simulation. We find that communication strategies learned using RL vary little from the director baseline strategies: they predict different behavior only across a small set of dialogue contexts for a limited range of simulated user profiles. In other words, users would have to turn out to have just the right behavior, and we'd have to collect voluminous data, in order to demonstrate a difference between the approaches in a user study. Intuitively, the system will always have little evidence for possible misunderstanding until the user follows up; once the user follows up, it doesn't take deep inference to provide a robust and effective clarification. 

We summarize the key conclusions from our analysis as follows:
\begin{itemize}
    \item A coherence-based collaborative approach is an attractive framework for system design when systems need to present and ground their own contributions.
    \item When systems support flexible interactive skills, such as answering clarification questions, simple strategies for deploying those skills may be hard to beat. 
\end{itemize}
These findings are in line with Clark and colleagues' \cite{CLARK19861} principle of least \emph{collaborative} effort---that human speakers' and audiences' strategies are simple and mutually responsive, rather than systematically optimized.  We are curious to explore this possibility in a wider range of conversational domains of practical interest.

\section{Related Work}
\label{sec:related}

There have been several efforts in the dialogue research literature to model initiative in situated conversations. In one influential prototype \cite{bohus2009open-world}, a receptionist system manages conversation initiative to interact with different customers in a situated conversation. Other conversational systems in task-oriented settings address information queries \cite{chu-carroll-2000-mimic} and collaborative problem solving \cite{Ferguson2005MixedInitiativeDS}. In such systems, task initiative is generally defined as requiring systems to guide the conversation so that the user specifies the task-specific parameters according to the system's expectations, and contrasted with mixed-initiative systems where the user specifies parameters more freely. User-initiated clarifications are not on the table, as confirmed by a survey of task-oriented conversational systems \cite{larsson-2017-user}, which reveals that fielded systems are gnerally incapable of answering clarifications by the user.

There is also work which aims to understand what role mixed-initiative plays in human--human interaction. The goal is to understand the dynamics of human communication to help in in building better conversation models. However these do not model the complex dynamics between the speakers involved in the conversation \cite{core-etal-2003-role, 10.1016/j.csl.2009.04.003, 10.1145/1753326.1753516}. Reinforcement learning (RL) has also been used to optimize communication policies over handcrafted baselines \cite{LEMON2011210, manuvinakurike-etal-2017-using, Misu2012ReinforcementLO}. For example, RL has been shown to enable adaptive and user-centric policies for initial information presentation \cite{janarthanam-lemon-2014-adaptive}. But such work has not addressed the rephrasing required to respond to user-initiated clarification.

\section{Problem Statement}
\label{sec:approach}

Here we will first provide a summary of the task and architecture of our dialogue system. We then move on to lay basic building blocks of how the model of system behavior induces a learning problem for communicating with a user. We then summarize the mechanisms behind the simulation models and the reinforcement learning frameworks we utilize to solve the resulting optimization problem.

\subsection{Colors in Context}
\label{cic_task}

\begin{figure}[bh]
    \centering
    \begin{tabular}{cc}
        \includegraphics[scale=0.33]{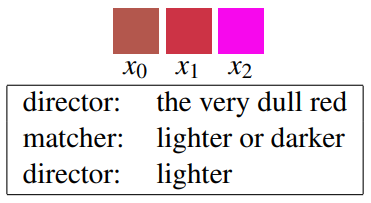} &
        \includegraphics[scale=0.47]{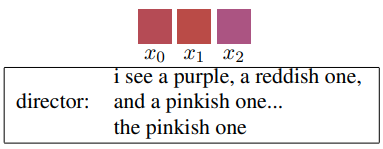}\\
    \end{tabular}    
    \caption{Figure shows two example interactions from CIC dataset.}
    \label{fig:cic_example}
\end{figure}

We use an established referential communication task  \cite{monroe-etal-2017-colors} to test the performance of our director model. The task involves showing participants, director and matcher, a set of three color patches $x_1$, $x_2$, $x_3$ in different permutations. The director knows which of the three patches is the target and has to identify the target to the matcher. The conversation data is collected in English through a text chat interface. A task example is shown the in Figure \ref{fig:cic_example}. 

The human--human conversation data is collected in three task difficulties: i) \textit{far} ii) \textit{split} iii) \textit{close}. The \textit{far} condition is easiest, since the color patches in this case generally come from different color categories. The \textit{split} condition has two color patches which look similar, while all color patches look similar in the \textit{close} condition, which makes it the hardest. Subjects sometimes find it hard to identify the target color patch in a single turn and the matcher regularly makes use of clarification questions to resolve any ambiguities in the director's explanations.

 Overall, human matchers are successful in selecting the correct target $\sim 90\%$ of the time. Around $\sim 97\%$ of the human conversations do not have clarifications: thed matcher selects the target just using the description in the first turn. Most of the other conversations conclude after a clarification question and a single director responds. This suggest that human directors are quite successful both in their initial descriptions and in their followup utterances.

\subsection{Director Communication Strategy}
Analyzing human-human conversations reveals a range of communication strategies that human directors use to describe the target color patch. For example, human directors sometimes make use of parallel descriptions for each color patch and then specify the target by repeating one description.  However, human description strategies can be formalized as a sequence of descriptions of individual referents. Consequently, we structure RL to learn a composition of different color patch descriptions so it will consider complex human-like communication strategies. We explain this in detail in Section \ref{sec:rl_setup}.

\subsection{Generating Color Patch Descriptions}

We use an existing cognitive model of color descriptions \cite{mcmahan-stone-2020-analyzing} to generate color descriptions. The model is based on the crowd-sourced collection of descriptions of color patches curated by Randall Monroe. The model offers several psychologically-plausible methods for effectively describing color patches which output a probability distribution $P(w_t | x_t, C)$, where $x_t$ is the target referent, $C$ is the context consisting of non-target color patches and $w_t$ is the color patch description at time $t$. We utilize the conservative speaker to generate color descriptions (its output is most reliable) and use the expectation maximization model to estimate the user's understanding (its inferences are most human-like).

\subsection{Approximating a State Posterior}

We utilize the coherence approach \cite{khalid-etal-2020-discourse} to model dialogue state. Each new utterance $w_t$ is first translated into a logical form which is obtained using a domain-specific NLU module. In this case, the module is a parser for a domain-specific probabilistic context-free grammar (P-CFG). The logical form is used to update the context state represented as a knowledge graph of coherence relations. The logical form adds a new node to the knowledge graph through coherence-based attachment and is resolved in context through the use of a cognitive model which translates it to a probability distribution $P(x_i | w_t)$. Since the logical form for an utterance $w_t$ attaches to a node representing a previous utterance $w_{t-1}$, a posterior $P(x_i | w_t, w_{t-1}, ... , w_1)$ can be obtained which summarizes the all the contributions in a chain of attachments $w_t, w_{t-1}, ... w_1$. 

\section{Reinforcement Learning Setup}
\label{sec:rl_setup}

We use deep q-learning algorithm (DQN) as the RL approach and specify all its necessary components here:

\subsection{State Vector}

As stated earlier, each new speaker contribution is attached into a knowledge graph of discourse relations to obtain an updated context representation. A probability distribution over the color patches $x_i$ is approximated using the cognitive models \cite{mcmahan-stone-2020-analyzing}. We serialize this knowledge graph into a vector to represent the state $s_t$ for the RL algorithm. Our state vector $s_t$ is given as:
\begin{multline*}
    s_t = \{P(x_i|w_1,..,w_t)\;\forall\;i; P(x_{target}|w_1,..,w_t); 
    a_1; a_2;...; a_n; d_{min}; d_{max}; d_{avg}; l_{conv}; pt\}
\end{multline*}
where $a_i \in A$ (set of matcher and director actions) indicates whether $i^{th}$ action has occurred previously, $d_{*}$ indicates the relevant distance between the color patches, $l_{conv}$ indicates the conversation length so far and $pt$ is a flag indicating whether the previous speaker was the matcher or the director. $pt$ is used by the RL model as an indicator of whether the director is continuing its turn. 

\subsection{Director Actions}

Our analysis reveals that human directors use a variety of creative communication strategies. The three most common strategies employed by human directors are i) $a$; ii) $a \& a*$; iii) $a \sim b*$; where $a$ and $a*$ represents two different descriptions for the target color patch and $b*$ represents the description for the distractor closest to the target. These strategies can be created through composition of basic color descriptions about target or distractor color patches. We structure RL such that the director agent keeps making decisions until it makes an \textit{end turn} decision so it can learn to compose color patch descriptions. \textit{Left} table \ref{tab:strat_summary} shows the basic color descriptions RL director can choose from.
Actions 3 and 5 in the \textit{left} table \ref{tab:strat_summary} are only used as a response to a matcher clarification. Composing actions 1 and 2 result in $a \& \sim (b | c)$ (called the extended referential strategy) results in a strategy which may be helpful in a \textit{close} difficulty case since it contrasts the target with the distractors providing additional signal for the matcher. Similary, composition of strategies 1 and 4 results in $a \& \sim b$ which is one of the strategies human directors utilize and is a relaxed version of the extended strategy.
\begin{table}[]
    \centering
    \begin{tabular}[width=0.45\linewidth]{c|l}
        & Basic Director Strategies\\
        \hline
        1 & a \\
        2 & $\sim$ b or c \\
        3 & Affirm a Clarification Term \\
        4 & Negate the Color Patch Closest to the Target \\
        5 & Negate a Clarification Term \\
        6 & End Turn
    \end{tabular}
    \begin{tabular}[width=0.45\linewidth]{c|l}
       Strategy  &  Definition\\
       \hline
       Direct & $a$ \\
       
       Extended & $a \& \sim (b | c)$ \\
       
       Mixed & Use \textit{extended} in close cases \\
             & otherwise used \textit{direct}.
    \end{tabular}
    \caption{i) Left table shows the basic description strategies used by the RL director to curate a target description. ii) Right table shows the logical forms for different director baseline policies.}
    \label{tab:strat_summary}
\end{table}


\subsection{Reward Function}

Drawing insights from the Paradise paradigm \cite{paradise-1997} we want the director to effectively describe the color patch in as few turns as possible. So, we formulate the reward function such that each new director term earns a penalty. Each task success earns a large reward and failure a small one. This formulation summarizes our reward function:
\begin{equation}
\label{eq:reward_func}
    reward = r_{outcome} + (r_{term} * term\_count)
\end{equation}
where $r_{outcome}$ specifies the reward for the task outcome, $r_{term}$ specifies the penalty for each new color description the director model uses and $term\_count$ is the number of color descriptions uses by the director.

\subsection{Matcher Simulation}
\label{matcher_sim_details}

To train an conversational agent it needs to interact with a companion so it can try out communication strategies and get the reward feedback it can learn from. Since interacting with humans is too expensive this is accomplished through human simulations \cite{shi-etal-2019-build, 10.1017/S0269888906000944, 4806280, https://doi.org/10.48550/arxiv.1612.05688, schatzmann-etal-2007-agenda}. For the analysis of learned RL policy we use two matcher simulations:
\begin{itemize}
    \item matcher always selects the color patch most likely to be the target.
    \item given a threshold, the matcher asks clarifications if the probability for the most-likely target is less than threshold.
\end{itemize}
Our analysis reveals that humans only ask clarifications around $3\%$ of the time. In addition, humans tend to ask clarifications about the two most-likely color patches most of the time. Due to this reason, we opt for the use of clarifications about the two most-likely color patches in the matcher simulation. To adjust the rate of clarifications by the matcher simulation, we adjust the select action threshold at $95\%$ such that this holds true. Since human interactions are noisy we also specify a small clarification error rate of $10\%$. The matcher asks problematic clarifications at this rate and these provide no signal regarding matcher's understanding of the conversation context. This allows DQN to learn conversation policy in a noisy setting.

\subsection{DQN Formulation}

We make use of deep q-learning (DQN) algorithm \cite{https://doi.org/10.48550/arxiv.1312.5602} to train the RL agent. The algorithm uses a policy $Q_\theta^p$ and a target network $Q_\theta^t$ to approximate the current and the future expected q-values respectively. Q-values for a given state and action are approximated using these two networks to compute the difference $\delta$:
\begin{equation}
\begin{aligned}
& \delta = Q_\theta^p(s_t, a_t) - 
 (r(s_t, a_t) + \gamma max_{a' \in A_{dir}}Q_\theta^t(s_{t+1}, a'))
\end{aligned}
\end{equation}
where $A_{dir}$ is the action set for the director, $\gamma$ is the discount factor and $r(s_t, a_t)$ is the reward for the action $a_t$ in the state $s_t$. The Adam optimizer is used to optimize the weights $Q_\theta^p$ and $Q_\theta^t$ such that $\delta^2$ is minimized. Similar to the traditional DQN, we make use of an experience replay memory to construct a dataset of state-action transitions compute the $\delta$ using sampled mini-batches from this replay memory.

\section{Baseline Formulation and Analysis}
\label{sec:baseline}

Relying on the insights drawn from human conversations, we formulate three director baseline policies. Formal notation for the director policies is shown in the \textit{right} table \ref{tab:strat_summary}.
\begin{itemize}
    \item first is a basic director which tries to identify the target without utilizing the distractor information in the task context---we call this the \emph{direct baseline} policy.
    \item  second policy resembles a director which is always extra careful and tries to provide extensive information to identify the target color patch---we call this the \emph{extended director} policy.
    \item third policy baseline involves using the extended policy for the \emph{close} condition and using the basic policy rest of the time---we call this the \emph{mixed director} policy.
\end{itemize}
Our analysis reveals that answering matcher clarifications bridges the performance gap between the direct and extended strategies. This means that RL model will be able to learn interesting communication strategies for the matcher who always selects the target given a description.
       
       

\subsection{Effect of Clarifications}

\begin{figure}
    \centering
    \begin{tabular}{cc}
         \includegraphics[width=0.49\linewidth]{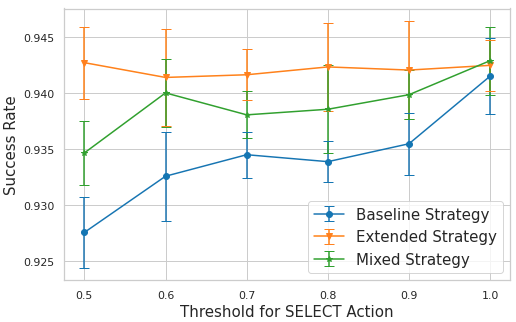} & 
         \includegraphics[width=0.49\linewidth]{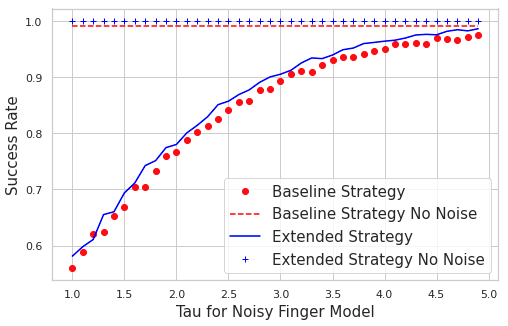}  \\ 
    \end{tabular}
    \caption{i) \textit{Left} figure shows the disadvantage of using ambiguous communication strategies vanishes as the user asks clarifications. ii) \textit{Right} figure shows the effect of parameter $\tau$ on the success rate for the direct baseline and extended strategies. Since the noise is added right before the select action each strategy is affected equally.}
    \label{fig:clar_task_outcome}
\end{figure}

To discern the room for flexible and context-sensitive director communication strategies we conduct a study on how changes in threshold for the \emph{select} action of the matcher affects the task outcome. At \textit{Left} in Figure \ref{fig:clar_task_outcome}, we show the effect of this change on task success. We find out that different strategies show a difference in their success rates when the user does not ask clarifications (threshold for the select action is low). However clarifications by the matcher diminish this difference which indicates that a rational matcher with the ability to clarify in case of ambiguity has the ability to make use of multiple ambiguous descriptions to arrive at the right answer. This also shows that RL agent will have the most room to learn trade-offs when it is dealing with a matcher who does not ask clarifications.

\subsection{Tuning Noise for a Realistic Setting}

Human-human conversations show a success rate of $\sim 90\%$ and human matchers use clarifications in minority of the cases $\sim 3\%$. As shown in the \textit{left} figure \ref{fig:clar_task_outcome} even when the threshold for select action is low the task success rate for the direct baseline strategy is $\sim 92\%$. One of the reasons for this high success rate is that there is no noise in the way we are evaluating probability distributions. Human actions in the real world are noisy so we use two noise inducing methods our matcher simulations.

\textbf{Noise Induction in Select Action}: We use a temperature based noise inducing parameter $\tau$ to perform a noisy softmax operation on the matcher's probability distribution and induce noise at the time of selection \cite{mcdowell-goodman-2019-learning}. We call this method the \textit{noisy finger} method. Lets use $p$ to represent the probability distribution $P(x_i | w_t, .., w_1)$ then noisy distribution $p_{tau}$ can be formulated as:
\begin{equation}
    p_{tau} =  softmax(\tau * p)
\end{equation}
At \textit{Right} in Figure \ref{fig:clar_task_outcome}, we show the effect of using $tau$ based noisy distribution on the success rates of both direct baseline and extended communication strategies. This method is successful for inducing noise because the temperature parameter $\tau$ affects the highest probability for the target disproportionately.

\begin{figure}
    \centering
    \begin{tabular}{c|c}
         \includegraphics[width=0.49\linewidth]{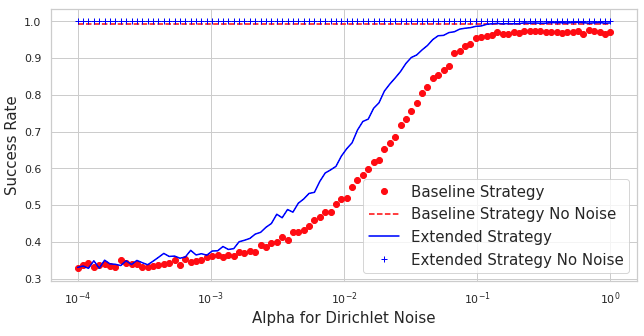} &
         \includegraphics[width=0.49\linewidth]{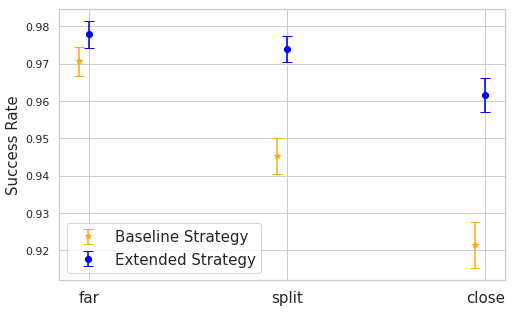}\\ 
    \end{tabular}
    \caption{i) \textit{Left} figure shows that $\alpha$ impacts both strategies differently. We can see that extended strategy outperforms the direct baseline across the board because it cumulates contributions of multiple descriptions. ii) \textit{Right} figures shows performance gain for extended strategy is observed in the \textit{close} task difficulty setting when $\tau = 4.5$ and $\alpha = 0.15$.}
    \label{fig:strat_differences}
\end{figure}

\textbf{Noise Induction in Semantic Interpretation}: To induce noise in the semantic interpretation of a director contribution we use a parameter $\alpha$ to sample a distribution from the gamma distribution obtained using the product of parameter $alpha$ and the probability distribution $p$ \cite{griffiths_canini_sanborn_navarro_2019}. Since the extended strategy involves composition of multiple contributions the noise is added to each probability distribution individually before obtaining the posterior. Thus, this operation affects the extended and direct baseline strategies differently. At \textit{Left} in Figure \ref{fig:strat_differences}, we show the impact of varying parameter $\alpha$ on the success rates of direct baseline and extended strategies. It reveals that due to combining information from multiple descriptions extended strategy is able to outperform the direct baseline strategy across the board. To make this concrete, for the noise parameter $\alpha=\sim 0.05$ where extended strategy achieves a success rate of $\sim 90\%$ the baseline (direct) strategy is only able to achieve a success rate of $\sim 75\%$

\subsection{Communication Strategy Choice Analysis}

Since the extended strategy shows an improvement over the direct baseline, we further analyze this improvement to better understand the impact of different communication strategies. Using the analysis presented above we adjust the noise inducing parameters $\tau = 4.5$ and $\alpha = 0.15$ such that each induces half of a realistic (human-human) error-rate. Our analysis of direct baseline and extended strategies, shown at \textit{Right} in Figure \ref{fig:strat_differences}, highlights that most of the performance gains occur in the \textit{close} task setting when utilizing the extended strategy. This suggests that RL should be able to learn strategies which improve the success rate for \textit{close} condition.


\subsection{Reward Function Analysis}

\begin{figure}
    \centering
    \begin{tabular}{c|c}
         \includegraphics[width=0.49\linewidth]{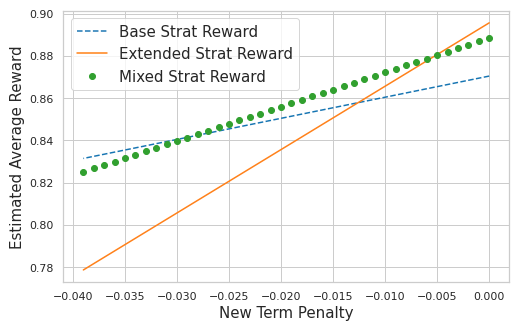} &
         \includegraphics[width=0.49\linewidth]{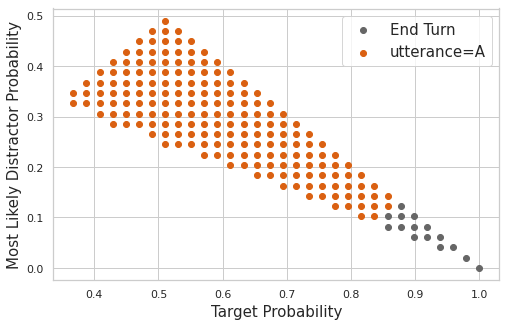}\\
    \end{tabular}
    \caption{i) \textit{Left} figure shows reward as a function of term use penalty when task success reward is $1.0$ and task failure reward is $-0.8$. ii) \textit{Right} figure shows the decisions of the DQN policy after an initial description as a function of the posterior probability of the target and the most-likely distractor. DQN chooses to end the turn when the target probability is above $\sim 84\%$.}
    \label{fig:policy_tradeoffs}
\end{figure}

The extended director strategy, though very thorough in structure, requires more effort from the director whereas the direct strategy does not utilize the external context information effectively. Our expectation for a DQN-based model is that it will learn a balance between some variations of direct and extended communication strategies. However, since DQN policies get their signals from a reward function we conduct a reward space analysis of the three director strategies specified above to identify the necessary parameters for the reward function specified in \ref{eq:reward_func} which will allow DQN to learn a flexible policy. At \textit{Left} in Figure \ref{fig:policy_tradeoffs}, we show the reward function for the three director hand-crafted policies as a function of penalty for the $term_count$ when $r_{success}=1.0$ and $r_{failure}=-0.8$.

Since we know the number of conversations for each difficulty we specify the $term\_count$ to be the average number of terms used by a director utilizing each policy. As depicted at \textit{left} in Figure \ref{fig:policy_tradeoffs}, there is a region in the space where mixed strategy achieves better performance in terms of the reward. Choosing a term penalty in this region will allow the DQN to learn a flexible policy.



\section{Analysis of a DQN Based Director Strategy}
\label{sec:expt}

In this section we first present the parameters we specified for the DQN learning process and then conduct a comparative analysis of the the hand-crafted communication strategies and those learned by the DQN algorithm. Our analysis suggests that DQN is able to learn a flexible communication strategy which outperforms the extended hand-crafted strategy in terms of the reward when interacting with the always selecting user but does not offer an advantage in terms of the success rate. When interacting with the clarifying matcher, DQN learns a variation of direct baseline strategy which is in line with our expectations, since answering clarifications bridges the performance gap between communication strategies.

In our experiments $Q_\theta^p$ and $Q_\theta^t$ are represented using a 2-layered dense network with a ReLU activation in between. The learning rate for the setting when: i) matcher model always selects the target is $10^{-2}$ and ii) matcher model asks clarification when appropriate is $7.5x10^{-5}$. Following from the reward analysis presented above we choose the penalty for additional color descriptions as $r_{term} = -0.025$. The noise parameters are $\tau = 4.5$ and $\alpha = 0.15$ to make the conversation setting realistic. We used 5000 CIC task contexts (set of three color patches) from the training set to generate simulation data for the experience replay memory used by the DQN algorithm. To test the policy we measure the success rate and the average reward of  using 1000 CIC contexts from the test set.

\subsection{Policy when the Matcher Always Selects}

In this setting the DQN model learns to describe the target color patch at the start of the conversation. The model proceeds to provide an additional description for the target if the probability of the target given the description is below the threshold of $\sim 84\%$. When the target posterior probability is above this threshold the DQN model procceeds to end the turn. Figure The \textit{right} figure \ref{fig:policy_tradeoffs} shows a visualization of the learned policy by the DQN model. 

The DQN policy outperforms the hand-crafted strategies by a slight margin in terms of the earned average reward which is in line with our reward space analysis presented above. A comparison of the learned policy and the hand-crafted director policies is shown in the table \ref{tab:dqn_perf_comp}. The DQN policy outperforms the direct policy in terms of the success rate but fails to outperform the extended strategy.

\begin{table}[]
    \centering
    \begin{tabular}{c|cc}
         Strategy &  Success Rate & Reward\\
         \hline
         DQN & $95.5\%$ & $0.891$\\
         Direct & $94.7\%$ & 0.880 \\
         Extended & $97.8\%$ & 0.874
    \end{tabular}
    \caption{This table presents a performance comparison between the handcrafted and learned director policies.}
    \label{tab:dqn_perf_comp}
\end{table}

\subsection{Policy when Matcher Clarifies Ambiguities}

As described in the section \ref{matcher_sim_details}, we specify additional parameters to tune the rate of clarifications and induce noise in the clarification questions. When interacting with this simulation, DQN model learns a variation of the direct policy which is indicative that it understands that clarifications diminish the advantage of using extended descriptions. The learned policy in this case has the following characteristics:
\begin{itemize}
    \item the DQN provides a target color patch description in the first turn.
    \item in case of a clarification the director responds with one of the terms matcher used to describe the target or by negating both the distractor color patches.
\end{itemize}
This shows that DQN agent understands it might have to re-describe the target if the probability distribution indicates that question is not referring the target. The DQN policy achieves a success rate of $95.9\%$ with an average reward of $0.901$ where the direct policy achieves a success rate of $95.8\%$ with a reward of $0.899$ policies.

\section{Discussion and Conclusive Remarks}

We present a detailed analysis of trade-offs when trying to learn a director model using a coherence based decision theoretic approach in a referential communication setting.  The coherence based state tracking approach outlined in \cite{khalid-etal-2020-discourse} coupled with RL is able to successfully learn flexible and context-sensitive communication strategies. However, our analysis reveals that a director which can answer clarification questions to resolve matcher ambiguities can bridge the performance gap between brief and detailed communication strategies. Because of these reasons using RL based techniques to learn context-specific communication policies is not practical when a simple director policy can get the job done. A detailed reward space analysis as presented above can help identify the utility of a RL approach.

In our evaluation it is revealed that strategies learned by the RL director and those crafted through analysis of human-human conversations are very similar. Any effect induced by these strategies with human subjects will too small to measure reliably with feasible experiment sizes, so we do not conduct human evaluations.

\subsection{Future Work and Conclusion}

One of the possible directions this work can go is to explore how our findings hold up in other domains e.g. slot-filling domains like restaurant booking \cite{budzianowski-etal-2018-multiwoz}. We hypothesize that since the job of a director is to guide the user to fulfill a certain task and answer any clarifications regarding task-specific parameters reliably, our obtained insights should sustain themselves in those domains. However we suspect this to be true for only those situations where initiative is held by the system. Many conversation scenarios could be mixed-initiative such that both the system and the user hold key pieces of information to complete a given task. In such a scenario, a model has to be able to answer clarifications reliably as well as clarify ambiguities. An example of such a scenario could be a conversation system deployed in a disaster control domain where job of the system is to guide the workers to help victims given the available information about the disaster site. In such a case the system will need to updates its understanding based on the new findings workers report about the disaster site e.g. a pile of rubble requires machinery to clear. This requires two-way communication about the world state and so involves different trade-offs.

Most dialogue research involves conversation models performing a reactive role where a user specifies the necessary parameters of interaction, as in e.g. a movie recommendation task. However, as these interfaces become more familiar and powerfl, people will utilize them for more complex tasks e.g. asking an automated agent to book a flight for them on phone. This requires the agent to take initiative in their interactions and ascertain uncertainty in user state so that they are able to answer clarifications effectively. In this paper, we present the challenges and trade-offs when trying to learn a communication policy in an environment where system holds the initiative. Our findings suggest that systems can get away with simple descriptions as long as they are able to answer clarification questions from the user effectively. In addition we find that empirical exploration of reward and action space is able to highlight the possible trade-offs and practicality for using RL.

\bibliographystyle{styles/spmpsci}
\bibliography{ms, anthology}
\end{document}